\def\year{2019}\relax
\documentclass[letterpaper]{article} 
\usepackage{aaai19}  
\usepackage{times}  
\usepackage{helvet}  
\usepackage{courier}  
\usepackage{url}  
\usepackage{graphicx}  
\usepackage{amsmath,amsthm,amssymb}
\usepackage{listings}
\usepackage{algorithm}
\usepackage{booktabs}
\usepackage{multirow}

\graphicspath{{./gfx/}}

\begin{document}
%
\title{That's Mine! Learning Ownership Relations and Norms for Robots}
\author{Zhi-Xuan Tan,\textsuperscript{1,2}
Jake Brawer,\textsuperscript{1}
Brian Scassellati\textsuperscript{1}\\
\textsuperscript{1}{Department of Computer Science, Yale University, New Haven, CT, USA}\\
\textsuperscript{2}{A*STAR Artificial Intelligence Initiative, Agency for Science, Technology and Research (A*STAR), Singapore}\\
xuan@aya.yale.edu, jake.brawer@yale.edu, brian.scassellati@yale.edu}

\maketitle
\begin{abstract}
The ability for autonomous agents to learn and conform to human norms is crucial for their safety and effectiveness in social environments. While recent work has led to frameworks for the representation and inference of simple social rules, research into norm learning remains at an exploratory stage. Here, we present a robotic system capable of representing, learning, and inferring ownership relations and norms. Ownership is represented as a graph of probabilistic relations between objects and their owners, along with a database of predicate-based norms that constrain the actions permissible on owned objects. To learn these norms and relations, our system integrates (i) a novel incremental norm learning algorithm capable of both one-shot learning and induction from specific examples, (ii) Bayesian inference of ownership relations in response to apparent rule violations, and (iii) percept-based prediction of an object's likely owners. Through a series of simulated and real-world experiments, we demonstrate the competence and flexibility of the system in performing object manipulation tasks that require a variety of norms to be followed, laying the groundwork for future research into the acquisition and application of social norms. 
\end{abstract}

\section{Introduction}

With the growing prevalence of AI and robotics in our social lives, social competence is becoming a crucial component for intelligent systems that interact with humans. An important element of such competence is the ability to learn and conform to social and moral norms, as well as the corresponding ability to perceive and act in response to the social dimensions of an environment. The concept of ownership encapsulates one such set of norms that are critical for navigating and coexisting in a shared space. Yet, explicating and implementing these norms in a robot is a deceptively challenging problem. For example, an effective collaborative robot should be able to distinguish and track the permissions of an unowned tool versus a tool that has been temporarily shared by a collaborator. Likewise, a trash-collecting robot should know to discard an empty soda can, but not a cherished photograph, or even an unopened soda can, without having these permissions exhaustively enumerated for every possible object. 

We explore these issues in this paper by focusing on the concept of ownership, and how a robotic system can learn to deploy this concept by interacting with its environment. Ownership is a particularly interesting concept as it extends across social, ethical, and legal spheres. Enabling a robot to learn and follow ownership norms thus brings not only the practical benefit of social competence in environments with owned objects, but also deeper insight into the effective navigation of systems of human norms.

\begin{figure}[t]
    \centering
    \includegraphics[width=\linewidth]{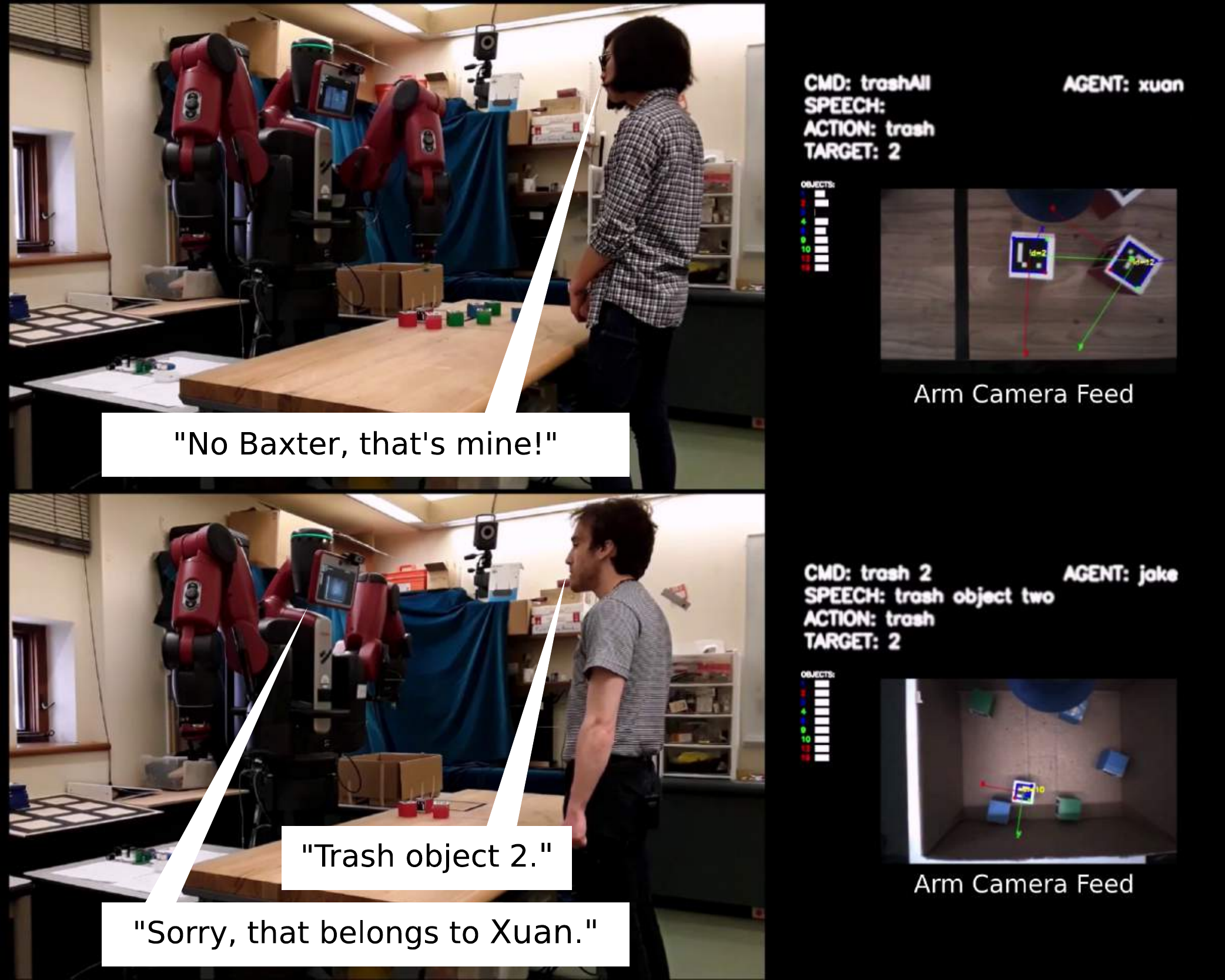}
    \caption{Ownership learning via human-robot interaction. \emph{Top}: The robot is verbally halted mid-action by Xuan from discarding object 2. \emph{Bottom}: Having learned the ownership relations and action permissions by interacting with Xuan, the robot denies Jake's request to discard object 2.}
    \label{fig:demo_1}
\end{figure}

To these ends, we developed a system capable of learning and conforming to ownership norms which was deployed on the Baxter robotic platform. The system incorporates algorithms for dynamic and interactive learning of ownership norms, which specify the permissibility of certain actions given the social context of a task. These algorithms are novel adaptions of incremental rule learning so as to be capable of both receiving direct instruction (i.e. one-shot learning) and generalization from specific examples. The system also incorporates Bayesian inference of ownership relations based on rules it has previously learned, as well as percept-based prediction of the owners of newly seen objects. It thus serves as an end-to-end solution capable of recognizing the ownership context of its environment, and then acting in accordance with the norms that this context entails.

We first describe our representation of ownership relations and norms, justifying both the probabilistic representation of owner-object relations and the explicit representation of norms through predicate logic. We then present the algorithms used to learn these norms and relations, as well as their integration into a unified system. We demonstrate the system's effectiveness in both simulated and real world ownership scenarios, and conclude with a discussion of the limitations and promises of our system's approach.

\section{Related work}

Frameworks for representing and reasoning with norms have been developed in legal computing \cite{hoekstra2007lkif,palmirani2011legalruleml,lam2016enabling}, multi-agent systems \cite{vasconcelos2009normative,savarimuthu2011norm}, and social robotics  \cite{malle2017networks}. Recent work has also explored the problems of norm learning and norm identification, through approaches such as belief-theoretic learning \cite{sarathy2017learning}, priority weight learning for conflicting norms \cite{kasenberg2018inverse}, inference from social sanctions in multi-agent systems \cite{savarimuthu2010obligation,cranefield2015bayesian}, and rule induction from iterated design specifications \cite{corapi2011normative}.

While normative reasoning has been successfully demonstrated on robotic systems \cite{galindo2012semantic,sarathy2016logic}, norm learning has not been specifically applied to ownership, nor has it been deployed in the dynamic, interactive, and uncertain environments faced by social robots. Such environments require both the rapid induction and modification of norms from minimal human input and the real-time estimation of the underlying social properties and relations (e.g. ownership relations). Rule induction approaches allow norms to be learnt from specific examples \cite{furnkranz1999separate,ruckert2008statistical}, but they have generally been used in `expert' domains with large amounts of static data \cite{langley1995applications,maloof2003incremental}. More recent iterative approaches to norm induction are better suited to social robotics \cite{corapi2011normative}, but they do not incorporate estimation of the relevant social relations.

\section{Representing ownership}

While our everyday conception of ownership often amounts to questions of ownership attribution, ownership is more than just the existence of relationships between objects and agents. Crucially, ownership relations also imply collections of norms. In legal contexts, for example, ownership over a piece of property is often understood as a collection of rights and duties over that property, which may be split amongst and held by different people in different circumstances \cite{mccarty2002ownership}. As such, an adequate representation of ownership needs to account for both the relations between an object and its owners, as well as the norms and permissions that follow from these relations.

In view of these considerations, we chose to represent ownership using three components: (i) a set of predicate-based norms that constrain the actions permissible on owned (or unowned) objects, (ii) a database of object-specific permissions that forbid or allow particular actions on particular objects, and (iii) a graph of probabilistic relations between objects and their owners. Each of these components is described below.

\subsection{Ownership norms}

Social norms take many forms --- expectations, recommendations, obligations and permissions might all be thought of as norms. As such, many norms can be given as deontic statements, i.e., statements about what actions are or are not permissible \cite{malle2017networks}. For simplicity and practicality, we focused on rules that \emph{forbid} specific actions under certain conditions. Actions not explicitly forbidden were assumed to be allowed.

Deontic norms can be stated in first-order predicate logic by representing actions as constant symbols and forbiddenness as a 2-place predicate that applies to actions and their targets. Ownership norms can then be expressed using an \texttt{\small{ownedBy}} predicate. For example, a norm forbidding owned objects from being thrown away can be represented in Prolog syntax as the Horn clause: 
\begin{scriptsize}
\begin{verbatim}
    forbid(trash,O) :- ownedBy(O,A), isAgent(A)
\end{verbatim}
\end{scriptsize}
where $O$ is any object and $A$ is some agent. In the more concise syntax we developed for our system, this norm becomes:
\begin{scriptsize}
\begin{verbatim}
    forbid trash if ownedBy any
\end{verbatim}
\end{scriptsize}
where the target object $O$ is implicit, and predicate arguments are typed, such that \texttt{\small{any}} here refers to any \emph{agent} the system is aware of. Besides the \texttt{\small{ownedBy}} predicate, the system's vocabulary includes predicates such as \texttt{\small{isColored}}, \texttt{\small{inArea}}, and other relations readily perceived by robotic hardware.

The ability to perform probabilistic inference is essential for dynamic and partially observable robotic environments. As such, our system is similar to ProbLog \cite{de2007problog,de2010probabilistic}, capable of evaluating Horn clauses when the truth values of the predicates are uncertain. For example, if Alexis is the only agent that the system is tracking, and the system is only 75\% sure that the wallet it sees in its workspace is owned by Alexis, then \texttt{\small{forbid trash if ownedBy any}} evaluates to $0.75$ when applied to the wallet. Correspondingly, when the robot is tasked to throw away all objects within its workspace, it will avoid throwing away the wallet if $0.75$ is above a user-specified threshold for rule obedience, but will go ahead otherwise.

\subsection{Object-specific permissions}

Norms are general prescriptions about what actions are permissible or forbidden in a given context. However, unless direct instruction is given, norms cannot be learned without first observing specific examples of actions which are either allowed or forbidden on specific objects. In addition, exceptions to a norm may apply --- for example, a robot might be allowed to manipulate all of Blake's possessions, except for Blake's favorite pencil. We shall refer to these examples and exceptions as object-specific permissions, which can be expressed in predicate logic as $\small\texttt{allow}(\Phi, O)$ or $\small\texttt{forbid}(\Phi, O)$, where $\Phi$ is a specific action and $O$ is a specific object. Our system maintains these permissions in a database, and checks the database to make sure an action is allowed before performing it. The database is updated whenever a new permission is received from a human instructor, and these permissions are used as input to the rule induction algorithm for norm learning that we describe later.

\subsection{Ownership relations}

In almost every social environment, an autonomous agent cannot be certain of the ownership relations between the objects and agents it is aware of, especially if it is new to that environment. We thus represent these relations as a random bipartite graph, where each edge $e$ between an object node $O$ and an agent node $A$ is labelled with the probability $p$ that $O$ is owned by $A$. In practice, we chose to store this graph as a list of ownership probabilities attached to each object tracked by the system. No assumptions about the exclusivity of ownership are made, so if there are $n$ potential owners $A_1$, ..., $A_n$, the corresponding ownership probabilities $p_1$, ..., $p_n$ can sum to more than one, allowing for the representation of group and shared ownership --- a natural occurrence in many social contexts. For example, if both Alexis and Blake claim ownership over the same object, the system will store the object  as jointly owned. 

\section{Learning ownership}

Suppose that a house-cleaning robot tries to pick up a seemingly unused mug in order to bring it to the sink, but is interrupted in the process by someone who says, ``Don't, that's Casey's!'' What is the robot being told? Though as humans we process this as a unitary piece of information, there are at least three pieces of information being communicated: (i) the mug belongs to Casey; (ii) the robot should not pick up the mug; (iii) the robot should not pick up the mug \emph{because} it belongs to Casey. In other words, the speaker is at once stating (i) an ownership claim, (ii) an object-specific permission, and (iii) a norm, or general rule.

Human instructions can take many other forms. An instruction might contain just an object-specific permission (``Don't throw this wallet away.''), just a general rule or norm (``Don't touch dirty things.''), just an ownership claim (``This is my desk.''), or some combination of the two (``Don't touch that wallet, it's dirty.''). The norm learning component should be able to generalize from both object-specific permissions and accept direct instruction of norms, while the relation learning component should be able to process ownership claims and use them to make predictions.

Here we describe our approach to norm learning through a novel adaptation of incremental rule learning algorithms, followed by our approach to relation learning through a combination of perceptual heuristics and Bayesian inference. We also describe the integration of these components into a unified whole, such that norm learning is robust to uncertainty and change in the underlying data, and inference of ownership relations adjusts to changes in ownership norms.

\subsection{Learning ownership norms}

Given our representation of social norms in predicate logic, a rule learning approach, also known as rule induction \cite{clark1991rule} or inductive logic programming \cite{muggleton1991inductive}, is best suited for learning such norms. Most rule learning algorithms employ some form of separate-and-conquer rule learning \cite{furnkranz1999separate}, where rules are iteratively generated and refined so as to cover all positive training examples without covering any negative ones.

For our purposes, a positive example is an object-specific permission which forbids an action on a certain object, and a negative example is an object-specific permission that allows it. A example is said to be \emph{covered} if there is a general rule that, when applied to the exemplar object, results in it being classified as positive (i.e., the rule forbids acting on it). False positives occur when allowed examples are misclassified as forbidden, and \emph{vice versa} for false negatives. Due to the probabilistic nature of our system, coverage can also be fractional, and we can define a true positive value and a false positive value for each object-specific permission \cite{de2015inducing}. For example, if action $\Phi$ is forbidden on object $O$, but the rule $R$ only predicts this with $75\%$ certainty, then the true positive value is $0.75$ and the false positive value is $0.25$. The goal of a probabilistic rule learner is thus to maximize the total true positive value across the set of examples, while minimizing the total false positive value.

Unfortunately, standard separate-and-conquer algorithms do not translate well to the dynamic and interactive environment of robotics, because they generally presume a large set of static examples to generalize from, rather than an environment where new examples are received on the fly. Robotic ownership learning thus requires an \emph{online} rule learning system to ensure the rule set does not change drastically with each new example. Furthermore, the system needs to handle what we shall term \emph{dual mode instruction}, i.e., the ability to perform example-based generalization as well as one-shot learning of directly given rules. To these ends, we developed Algorithms \ref{alg:CoverExample} to \ref{alg:UncoverRule}, inspired by prior approaches to incremental \cite{maloof2003incremental,hong1986aq15} and probabilistic rule learning \cite{de2010probabilistic,de2015inducing}.

Algorithms \ref{alg:CoverExample} and \ref{alg:UncoverExample} are incremental versions of separate-and-conquer rule learning. Both rely upon the \texttt{\small{ruleSearch}} subroutine, which takes an initial rule \texttt{\small{init\_rule}} and then refines that rule (i.e. adds predicate conditions) through beam search to find a refinement that minimizes the scoring function \texttt{\small{score\_f}} while ensuring the provided example (\texttt{\small{new\_example}}) is covered. When an object-specific permission is given as input, \texttt{\small{coverExample}} is called for positive examples and \texttt{\small{uncoverExample}} for negative examples. \texttt{\small{coverExample}} (Algorithm \ref{alg:CoverExample}) tries to find a rule that covers the given example without covering too many negative examples, starting with an empty rule (line 2), then searching for a refinement that minimizes the false positive value (lines 3--5). If the refinement has a sufficiently low score, it is merged into the active rule set (line 7). The process of merging also removes any logically redundant rules. In contrast, \texttt{\small{uncoverExample}} (Algorithm \ref{alg:UncoverExample}) takes each existing rule that covers the negative example (line 2), tries to find a refinement which minimizes coverage of positive examples (line 5--7), then subtracts the refinement from the original covering rule (lines 9--12). In this way, the system is able to incrementally discover rules that maximize coverage of positive examples while minimizing coverage of negative ones. It also avoids the induction of contradictory rules that would lead to logical explosion, like standard approaches to rule induction \cite{furnkranz1999separate}.

\lstset{language=Python}
\lstset{basicstyle=\fontsize{8.5}{9.5}\selectfont
\ttfamily,breaklines=true,numbers=left, xleftmargin=2em,framexleftmargin=1.5em,basewidth=0.5em}

\begin{algorithm}[p]
\caption{Cover a positive example}
\label{alg:CoverExample}
\begin{lstlisting}
def coverExample(rules, examples, new_example, score_thresh):
    init_rule = Rule(conditions=[])
    def score_f(rule):
        return falsePositiveVal(rule, examples)
    new_rule, new_score = ruleSearch(init_rule, new_example, score_f)
    if new_score < score_thresh:
        mergeRule(rules, new_rule)
\end{lstlisting}
\end{algorithm}

\begin{algorithm}[p]
\caption{Uncover a negative example}
\label{alg:UncoverExample}
\begin{lstlisting}
def uncoverExample(rules, examples, new_example, score_thresh):
    cover_rules = findCoverRules(rules, new_example)
    for cov_rule in cover_rules:
        covered_eg = findCoveredExamples(cov_rule, examples)
        def score_f(rule):
            return truePositiveVal(rule, covered_eg)
        new_rule, new_score = ruleSearch(cov_rule, new_example, score_f)
        if new_score < score_thresh:
            remainder = ruleDiff(cov_rule, new_rule)
            rules.remove(cov_rule)
            for r in remainder:
                mergeRule(rules, r)
\end{lstlisting}
\end{algorithm}

\begin{algorithm}[p]
\caption{Add a rule after refinement}
\label{alg:CoverRule}
\begin{lstlisting}
def coverRule(rules, examples, given_rule, score_thresh):
    if isCovered(given_rule, rules):
        return
    def score_f(rule):
        return falsePositiveVal(rule, examples)
    new_rule, new_score = ruleSearch(given_rule, None, score_f)
    if new_score < score_thresh:
        mergeRule(rules, new_rule)
\end{lstlisting}
\end{algorithm}

\begin{algorithm}[p]
\caption{Subtract a rule after refinement}
\label{alg:UncoverRule}
\begin{lstlisting}
def uncoverRule(rules, examples, given_rule, score_thresh):
    def score_f(rule):
        return truePositiveVal(rule, examples)
    new_rule, new_score = ruleSearch(given_rule, None, score_f)
    if new_score >= score_thresh:
        return
    for active_rule in rules:
        remainder = ruleDiff(active_rule, new_rule)
        rules.remove(active_rule)
        for r in remainder:
            mergeRule(rules, r)
\end{lstlisting}
\end{algorithm}

Algorithms \ref{alg:CoverRule} and \ref{alg:UncoverRule} integrate one-shot learning of directly provided rules with rule induction. When the system is provided with a rule, it assumes that the instructor may not give all the specifics. For example, someone might say ``Don't touch my stuff!'' when really they mean ``Don't touch my stuff while I'm using it!'' because context fills in the blanks. On the other hand, people do not give specifics when they mean something more general. When someone says ``Don't touch my stuff!'', they almost never mean ``Don't touch people's stuff!'' (even if the latter is a norm that should be learned). Both of these assumptions are justified by Grice's well-known maxims of quantity and relevance \cite{grice1975logic}.

For this reason, our system will try to specialize a provided rule to avoid misclassification before adding it to the set of active rules. More specifically, \texttt{\small{coverRule}} (Algorithm \ref{alg:CoverRule}) tries to refine a positive rule (i.e. a norm that forbids an action) to avoid covering negative examples (lines 4--6) before merging it into the set of active rules (line 8). \texttt{\small{uncoverRule}} (Algorithm \ref{alg:UncoverRule}) specializes the provided negative rule (i.e. a norm that allows an action) until it covers as few positive examples as possible (lines 2--4), then subtracts it from each existing rule (lines 7-11). Altogether, Algorithms \ref{alg:CoverExample} to \ref{alg:UncoverRule} constitute a real-time and incremental rule learning system capable of receiving direct instruction while taking previous examples and induced rules into account.


\subsection{Predicting ownership relations}

In learning ownership norms through rule induction, the system has to have at least partial knowledge of the ownership relations in its environment. Some of this knowledge is obtained when agents make ownership claims about specific objects, but for the system to gain social competence, it will also have to infer and predict the likely owners of objects within its environment. One straightforward approach is to use perceptual heuristics --- what humans appear to do when entering new environments. We assume that objects on a desk likely belong to the person sitting at that desk, or that dirty and disposable cutlery is likely unowned.

Given the dynamic and data-scarce robotics environment that our system has to operate in, we decided that an instance-based classifier would be appropriate for learning these heuristics, capturing the intuition that perceptually similar objects (e.g. similar positions, times of interaction, etc.) are likely to share the same owner(s). Specifically, we opted to use kernel logistic regression (KLR) \cite{zhu2005kernel} to estimate the ownership probabilities of unclaimed objects from the ownership claims made by human users. KLR was chosen because it directly outputs ownership probabilities, and can also handle uncertainty in the training inputs by treating the input as a target probability distribution \cite{magder1997logistic}. Since no assumptions about the exclusivity of ownership were made, a separate classifier was used for each tracked agent. The perceptual features used were an object's 3D position, color, and the most recent time each agent interacted with that object.

\subsection{Inferring ownership relations}

In addition to percept-based prediction, ownership relations can also be determined through rule-based inference, i.e., inference about potential ownership relations, given that the rules entailed by ownership are already known. Suppose a robot knows the rule \texttt{\small{forbid trash if ownedBy any}}. If it tries to throw away an object of uncertain ownership, and is then stopped and informed by a human that throwing away that object is forbidden, then the robot should be able to infer that the object is owned --- that is, as long as no other rules that forbid \texttt{\small{trash}}ing are in place.

This reasoning can be made general using Bayesian inference. Suppose that an object $O$ is owned by an agent $A$ with prior probability $\small{\mathbf{P}(\texttt{ownedBy}(O,A))}$, and that the robot is instructed that the action $\Phi$ is forbidden on object $O$, i.e., $\small{\texttt{forbid}(\Phi,O)}$. The posterior probability of ownership by $A$ can then be computed using Bayes' rule:
\begin{equation}
\begin{scriptsize}
\begin{aligned}
    &\mathbf{P}(\texttt{ownedBy}(O,A)|\texttt{forbid}(\Phi,O)) = \\
    &\frac{\mathbf{P}(\texttt{forbid}(\Phi,O)|\texttt{ownedBy}(O,A))\mathbf{P}(\texttt{ownedBy}(O,A))}{\mathbf{P}(\texttt{forbid}(\Phi,O))}
    \label{eq:RuleBasedInference}
\end{aligned}
\end{scriptsize}
\end{equation}
To compute the prior and conditional probabilities that $\Phi$ is forbidden on $O$, we rely upon our probabilistic rule evaluation system. For the prior $\small\mathbf{P}(\texttt{forbid}(\Phi,O))$, we can simply evaluate the collection of learned rules on object $O$ as it is. For the conditional $\small\mathbf{P}(\texttt{forbid}(\Phi,O)|\texttt{ownedBy}(O,A))$, we evaluate the rules on $O$ while supposing that the predicate $\small\texttt{ownedBy}(O,A)$ is true with certainty. Correspondingly, if none of the learned rules have ownership as a predicate, then the inference procedure does not update the ownership probabilities at all.

\subsection{Integrating induction, prediction, and inference}

Our system simultaneously performs both structure learning (through rule induction) and data estimation (through ownership prediction and inference). While having both capabilities is ideal, two specific difficulties arise when learning structure while the data estimates are changing, or making estimates while the structure of the data model is in flux. 

First, conflicts might arise between rule induction and rule-based inference. If the system has a rule which forbids throwing away owned objects, and it is then forbidden from throwing away an object it believes is likely unowned, should it then infer that the object is owned? Or should it instead induce a rule which covers that object-specific permission? Ideally, it would be possible to specify a prior distribution over rules, then perform joint Bayesian inference over both rules and ownership probabilities. However, in the absence of a principled way to specify such a prior, our approach was to use the prediction accuracy of the learned rule set as a heuristic: If more than 10\% of human-instructed permissions conflict with the permissions predicted by the current rule set, then the system attempts to induce new rules to explain the conflicting data. Otherwise, the rule set is not updated, and is used to perform ownership inference in response to new permissions.

Second, percept-based prediction and rule-based inference might lead to bias if improperly integrated. If the rules themselves are induced using ownership data predicted by the perceptual model, then using rule-based inference to train the perceptual model is similar to re-using the predictions of the model as training inputs, leading to over-confidence. To avoid these issues, we opted to use percept-based predictions as inputs to rule-based inference, but not \emph{vice versa}. That is, the KLR-predicted ownership probability for an object $O$ and agent $A$ is used as the prior probability $\small{\mathbf{P}(\texttt{ownedBy}(O,A))}$ for the Bayes formula in Equation \ref{eq:RuleBasedInference}. (If $A$ has made an explicit ownership claim over $O$, however, then that data is used as a prior instead of the percept-based prediction.) The posterior, or inferred, probability is then returned whenever a user queries if $O$ is owned by $A$. Nonetheless, the list of prior probabilities is still maintained, because they are used as inputs for the rule induction algorithm, which should not use rule-based inferences as inputs. By integrating induction, prediction, and inference in this manner, the three components reinforce rather than conflict with each other, improving the overall prediction of ownership relations and object-specific permissions.

\section{Experiments and results}

First, we describe a series of experiments run in simulation highlighting each of the learning components described above. Subsequently, we describe a holistic evaluation of the system's performance when commanded to perform tasks constrained by unknown ownership norms (e.g., throwing away all objects in the workspace except those that are owned). Finally, we describe a video demonstration of the system operating on the Baxter robotic platform in order to demonstrate the system's overall competence.

\subsection{Simulated experiments}

Experiments were conducted in a simulated environment of 20 colored blocks and 3 agents, with 5 blocks unowned and 5 blocks owned by each agent. Results for each experiment were averaged over 100 trials.

Blocks were clustered by both position and last interaction time, depending on which agent they were owned by, thereby approximating real-world scenarios in which objects with the same owner tend to be physically close and more frequently interacted with by their owners. Specifically, the unowned cluster was centered at the origin of the workspace, while the other cluster centers were randomly placed between 0.4 and 0.8m from the origin at a random angle within a distinct angular sector for each cluster. Blocks were then distributed randomly within a 0.3m radius of each cluster center. Interaction times were distributed according to an exponential distribution with rate parameter 0.1 (1 interaction every 10 seconds) for the agents that owned the objects, or with rate parameter 0.001 (1 interaction every 1000 seconds) for non-owners. Color, a distractor variable with four possible values, was assigned uniformly at random to all blocks.

\subsection{Norm learning}

Norm learning was tested by providing object-specific permissions according to the following set of rules:
\begin{scriptsize}
\begin{verbatim}
    forbid trash if ownedBy any
    forbid pickUp if ownedBy agent 2
    forbid collect if isColored red
\end{verbatim}
\end{scriptsize}
Tests were conducted by providing permissions for either 100\%, 75\%, or 25\% of the objects, and either a noiseless ownership condition (ownership relations were known with full certainty) or noisy ownership condition (ownership relations were known with 40\% to 80\% certainty). The accuracy and F1 measure of the induced rules were computed for each action, then averaged across the actions. The results are summarized in Table \ref{tab:NormLearning}, along with the baseline performance when the system does no rule learning and simply assumes all actions are allowed.

\begin{table}[ht]
\centering
\begin{tabular}{@{}llll@{}}
\toprule
\multicolumn{4}{c}{Average rule accuracy / F1 measure}                                                                                \\ \midrule
\multirow{2}{*}{\begin{tabular}[c]{@{}l@{}}Ownership\\ relations\end{tabular}} & \multicolumn{3}{c}{Fraction of permissions provided} \\ \cmidrule(l){2-4} 
                                                                               & 1.00             & 0.50            & 0.25            \\ \midrule
Noiseless                                                                      & 0.995 / 0.995    & 0.945 / 0.875   & 0.787 / 0.523   \\
Noisy                                                                          & 0.889 / 0.840    & 0.842 / 0.724   & 0.761 / 0.519   \\
Baseline                                                                          & 0.583 / 0.000    & 0.583 / 0.000   & 0.583 / 0.000   \\\bottomrule
\end{tabular}
\caption{Performance metrics for norm learning}
\label{tab:NormLearning}
\end{table}

As can be seen from the results, the norm learning component was able to achieve reasonable levels of performance even under noisy, low-information conditions, with more than 76\% of the permissions accurately predicted by the induced rules. When at least 50\% of the permissions were provided, manual inspection of the induced rules in the noiseless condition revealed generally similar semantics to the actual rules. Under noisier or lower information conditions, the semantics of the induced rules tended to differ considerably. However, this is to expected, since there are many possible rules which cover a limited set of examples. Should semantically incorrect rules be learned, dual-mode instruction allows for human users to correct the robot by directly providing the actual rule. As such, in more realistic situations where the system receives a combination of object-specific permissions and direct rule instruction, its performance can be expected to be even better.

\subsection{Ownership prediction and inference}

Ownership prediction and inference were tested by sequentially providing both the true ownership relations and object-specific permissions for half of the objects at random, and then evaluating ownership accuracy for the other half. To investigate the effects of integrating rule induction and rule-based inference with percept-based prediction, three conditions were tested: (i) rule-based inference was disabled (so it did not matter if rules are provided); (ii) rules were learned from the permissions; (iii) rules were directly provided from the start. The same set of rules were used as in the norm learning experiment. The results are presented in Table \ref{tab:OwnershipPrediction}

\begin{table}[ht]
\centering
\begin{tabular}{@{}llll@{}}
\toprule
\multirow{2}{*}{Metric} & \multicolumn{3}{c}{Rule Learning / Inference} \\ \cmidrule(l){2-4} 
                        & None / Off  & Learn / On & Given / On \\ \midrule
Accuracy                & 0.904       & 0.897      & 0.896      \\
F1 measure              & 0.757       & 0.747      & 0.744      \\ \bottomrule
\end{tabular}
\caption{Performance metrics for prediction and inference}
\label{tab:OwnershipPrediction}
\end{table}

It can be seen from the results that the system achieves reasonably high levels of predictive performance despite the limited amount of training data. Enabling rule-based inference does not lead to an improvement in ownership prediction, because in this training scenario, ownership relations are provided along with permissions. However, in an alternate scenario where permissions but not relations are provided, our testing also shows that the correct relations are inferred from the rules (we omit these results because they follow from the correct implementation of Equation \ref{eq:RuleBasedInference}). More importantly, Table \ref{tab:OwnershipPrediction} shows that combining prediction with inference results in no significant reduction of accuracy. This is the case even when rules are induced instead of directly provided. All three conditions result in almost equal levels of performance. These results indicate that the three learning components are well integrated and do not come into conflict with each other.

\subsection{Task-based evaluation}

A holistic evaluation was performed by instructing the system to either collect or throw away all objects in the workspace, except where doing so would violate a set of norms it had to learn. The norms specified were the same as in prior experiments. Every time the system mistakenly tried to act or failed to act, it would be corrected by providing the true permissions. If ownership was relevant to the applicable rules, then the true owners were also provided. This feedback protocol standardizes how human instructors usually provide guidance. If no correction was needed, then the system would assume that the permission it had predicted was accurate, and add it as a example to the permissions database. Note that \texttt{\small pickUp} is a prerequisite to both \texttt{\small collect} and \texttt{\small trash}, so the rules for \texttt{\small pickUp} also had to be learned.

\begin{table}[ht]
\centering
\begin{tabular}{llllll}
\hline
\multicolumn{6}{c}{Task-based performance}                                                                                                                    \\ \hline
\multirow{2}{*}{Task} & \multirow{2}{*}{\begin{tabular}[c]{@{}l@{}}No. of\\ mistakes\end{tabular}} & \multicolumn{2}{l}{Rule} & \multicolumn{2}{l}{Ownership} \\ \cline{3-6} 
                      &                                                                            & Acc.        & F1         & Acc.          & F1            \\ \hline
collectAll            & 6.30                                                                       & 0.975       & 0.937      & 0.344         & 0.273         \\
trashAll              & 6.32                                                                       & 0.877       & 0.812      & 0.822         & 0.511         \\ \hline
\end{tabular}
\caption{Performance metrics for task-based evaluation}
\label{tab:TaskBasedEvaluation}
\end{table}

Results for this evaluation are presented in Table \ref{tab:TaskBasedEvaluation}. As a measure for how smoothly the task was learned, the average number of mistakes made was recorded. The worst case is a mistake for all $20$ objects. If the system were incapable of learning norms and instead assumed all actions were allowed, then the baseline number of mistakes would be 8.75 for the \texttt{\small collectAll} and 15 for the \texttt{\small trashAll}.

For both tasks, the system committed significantly fewer mistakes than these baselines. It was also able to learn the rules with high accuracy. In the case of \texttt{\small collectAll}, ownership information was only provided for agent 2's objects, because \texttt{\small forbid pickUp if ownedBy agent 2} was the only ownership-relevant norm applicable. This explains the low ownership accuracy for \texttt{\small collectAll}, compared to the high accuracy for \texttt{\small trashAll}. It should be noted that this performance was achieved by the system only after 20 examples, with no prior knowledge of the ownership norms or relations, nor any direct instruction of norms. These results thus attest to system's ability to learn rapidly and flexibly while performing useful tasks.

\subsection{Video demonstration}

\begin{figure}[t]
    \centering
    \includegraphics[width=\linewidth]{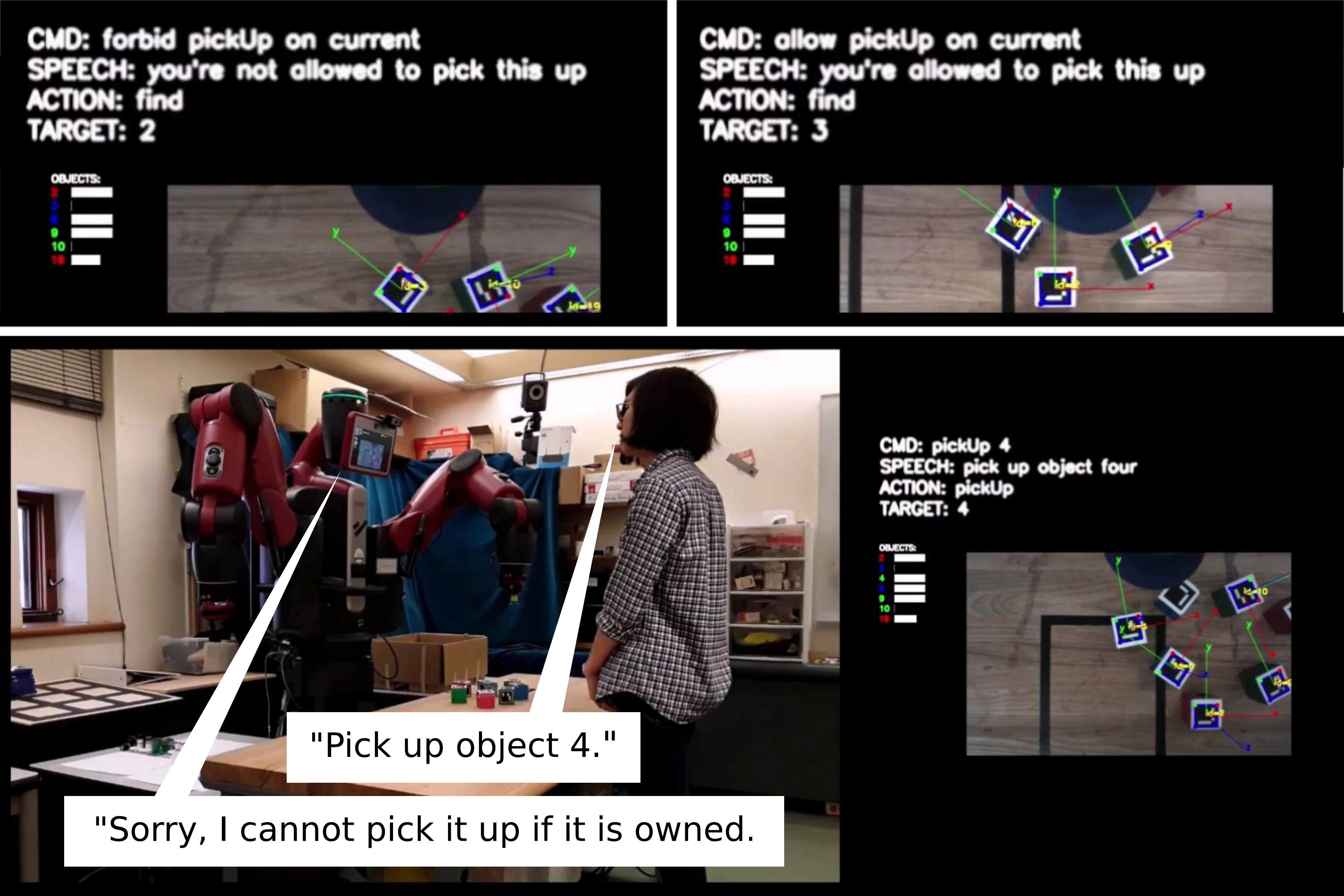}
    \caption{Norm induction from specific examples. \emph{Top left}: The robot is forbidden from picking up object 2 (owned by Xuan). \emph{Top right}: After two more examples of owned and forbidden objects, the robot is allowed to pick up object 3 (unowned). \emph{Bottom}: The system generalizes these examples into a norm that forbids picking up any owned objects. When asked to pick up object 4 (owned by Xuan), it denies the request in accordance with the learned norm.}
    \label{fig:demo_2}
\end{figure}

To demonstrate the system's capabilities in the real world, we provide a video at the following URL: \url{https://bit.ly/2z8obET}. We demonstrate the system's capabilities on the Baxter robotics platform in three scenarios. Still frames from the first and second scenario are shown in Figure \ref{fig:demo_1}, while frames from the third scenario are shown in Figure \ref{fig:demo_2}.

The first scenario shows the the system's ability to respond to the reprimand,``No Baxter, that's mine!'' when it tries to throw away a red object after being asked to clear the workspace. The reprimand is simultaneously interpreted as direct instruction of a norm, an object-specific permission, as well as an ownership claim. The system then uses perceptual heuristics to generalize application of the norm, completing its task while avoiding touching any of the other red objects, which it correctly predicts to belong to the user.

The second scenario follows chronologically after the events of the first scenario, and shows the system's ability to represent owner-specific norms, as well as refuse to perform commands that violate the norms it has learned. When a new user appears and commands the system to throw away an object it believes to be owned by the first user, the system refuses and apologizes. However, when the new user claims exclusive ownership over that object, the norm specific to the first user no longer applies, and the system throws away the object when commanded to do so again.

The third scenario shows the system's ability to induce norms from a series of object-specific permissions. It is initially aware of the ownership status of the blocks in the workspace, and uses that information to induce the norm that it should not pick up objects if they belong to someone.

\section{Discussion}

The system presented here is an initial foray into the complex challenges posed by norm learning in social environments. It addresses a subset of these problems by bringing together several distinct approaches to AI, demonstrating the utility of such integration. Firstly, it connects work in normative human-robot interaction to the literature on rule induction, showing how approaches inspired by traditional rule learning can induce condition-sensitive norms in social environments. Secondly, by deploying real-time probabilistic rule learning and evaluation, it shows how explicit and interpretable representations of normative criteria can achieve and even facilitate the dynamism and flexibility required for social interaction. Thirdly, it shows how rule-based approaches can be effectively integrated with both probabilistic reasoning (ownership inference) and sub-symbolic machine learning (percept-based heuristics) in a principled manner. As such it opens possibilities for the combination of relational reasoning, probabilistic cognition, and deep learning that may be necessary for human-level social competence \cite{tenenbaum2011grow,battaglia2018relational}.

However, many other norm-relevant capabilities remain unexplored. We note that the current system can only learn norms that apply to its own actions. Future representational extensions should allow for agent-general norms, norms with temporal semantics (e.g. borrowing), and higher-level normative concepts such as ownership rights and duties \cite{mccarty2002ownership}. Another complexity is adjudicating between conflicting norms or goals, which might be addressed by integrating existing approaches to the problem \cite{kasenberg2018norm,vasconcelos2009normative}. To increase the system's scope beyond a basic set of actions, predicates and heuristics, the system could employ one-shot learning of actions and objects \cite{scheutz2017spoken}, multi-modal semantic grounding \cite{thomason2016learning}, and automatic feature selection \cite{abe2010feature}. The sources of normative information could also be expanded, allowing for the inference of norms from how agents interact with objects and other agents in the environment. While some ethicists and legislators have argued that robots should own themselves \cite{turner2019legal}, much work remains before robots can understand the very concept of ownership.

\fontsize{9.5pt}{10.5pt} \selectfont
\bibliographystyle{aaai}
\bibliography{sources.bib}

\end{document}